\documentclass[9pt,twocolumn,twoside]{gsajnl}
\articletype{inv} 
\usepackage{lmodern}
\usepackage{amsmath,amssymb,amsfonts}

\usepackage{caption,subcaption}
\usepackage[para]{footmisc}
\usepackage{booktabs}
\usepackage{rotating}
\usepackage{wrapfig}
\usepackage{listings}
\usepackage{lipsum}
\usepackage{verbatim}
\usepackage{hyperref}
\usepackage{cleveref}

\runningtitle{Karim and Decker \textit{et al.}, Large Language Models to Knowledge Graphs} 
\runningauthor{Karim and Decker \textit{et al.}}

\title{From Large Language Models to Knowledge Graphs for Biomarker Discovery in Cancer}

\author[1$\dagger$]{Md. Rezaul Karim}
\author[2]{Lina Molinas Comet}
\author[3,2]{Md Shajalal}
\author[4,2]{Oya Deniz Beyan}
\author[5,6]{Dietrich Rebholz-Schuhmann}
\author[1,2]{Stefan Decker}

\affil[1] {Information Systems and Databases, RWTH Aachen University, Germany }
\affil[2] {Department of Data Science and Artificial Intelligence, Fraunhofer FIT, Germany }
\affil[3] {University of Siegen, Germany}
\affil[4] {University of Cologne, Faculty of Medicine and University Hospital Cologne, Institute for Biomedical Informatics, Germany}
\affil[5] {ZBMED - Information Center for Life Sciences, Germany}
\affil[6] {Faculty of Medicine, University of Cologne, Germany}

\correspondingauthoraffiliation[$\ast$]{\textbf{Corresponding author}:~\url{rezaul.karim@rwth-aachen.de}. This paper has been accepted in the ``Technical, Socio-Economic, and Ethical Aspects of AI'' research track of the 57th Hawaii International Conference on System Sciences~(HICSS'2024).}

\begin{abstract}
    Domain experts often rely on most recent knowledge for apprehending and disseminating specific biological processes that help them design strategies for developing prevention and therapeutic decision-making in various disease scenarios.~A challenging scenarios for artificial intelligence~(AI) is using biomedical data~(e.g., texts, imaging, omics, and clinical) to provide diagnosis and treatment recommendations for cancerous conditions.~Data and knowledge about biomedical entities like cancer, drugs, genes, proteins, and their mechanism is spread across structured (knowledge bases~(KBs)) and unstructured (e.g., scientific articles) sources.~A large-scale knowledge graph~(KG) can be constructed by integrating and extracting facts about semantically interrelated entities and relations.~Such a KG not only allows exploration and question answering~(QA) but also enables domain experts to deduce new knowledge. However, exploring and querying large-scale KGs is tedious for non-domain users due to their lack of understanding of the data assets and semantic technologies.~In this paper, we develop a domain KG to leverage cancer-specific biomarker discovery and interactive QA. For this, we constructed a domain ontology called OncoNet Ontology~(ONO), which enables semantic reasoning for validating gene-disease~(different types of cancer) relations. The KG is further enriched by harmonizing the ONO, metadata, controlled vocabularies, and biomedical concepts from scientific articles by employing BioBERT- and SciBERT-based information extractors.~Further, since the biomedical domain is evolving, where new findings often replace old ones, without having access to up-to-date scientific findings, there is a high chance an AI system exhibits concept drift while providing diagnosis and treatment.~Therefore, we finetune the KG using large language models~(LLMs) based on more recent articles and KBs.
\end{abstract}


\dates{\rec{xx xx, xxxx} \acc{xx xx, xxxx}}

\begin{document}

\maketitle
\thispagestyle{firststyle}
\vspace{-13pt}

\section{Introduction}
 A challenging scenario for AI is providing improved diagnosis and treatment for cancerous conditions~\cite{ballester2021artificial,tran2021deep}. With more than 200 different types identified, cancer is the second leading cause of death worldwide. Therefore, early detection and diagnosis, personalized interventions, and identifying biomarkers and therapeutic targets are of utmost importance in cancer~\cite{ballester2021artificial}. Cancer is characterized as a heterogeneous disease, having many types and subtypes. It is caused when cells turn abnormal, divide rapidly, and spread to other tissues and organs and may be further driven by a series of genetic mutations of genes induced by selection pressures of carcinogenesis in the cells. Further, the so-called marker genes including oncogenes and tumor suppressor genes are often responsible for cancer growth. 

Machine learning~(ML) techniques are widely used for the analysis of multimodal data~(e.g., multi-omics, texts, imaging, disease progression, etc.). Clinical data can also be combined into and analyzed using ML models~\cite{kitsios2023recent}. 
Further, cancer research requires datasets often generated from omics technologies and electronic health records. Domain experts often need to rely on up-to-date findings from a vast array of knowledge about drugs, genes, protein, and their mechanisms that are spread across structured~(knowledge bases~(KBs)) and unstructured data sources~\cite{karim2022explainable}. Scientific literature is a huge source of unstructured sources of biological entities. However, since these data\footnote{\scriptsize{For example, PubMed contains millions of articles~\cite{xu2020building}.}} are mostly unstructured, it makes the knowledge extraction overly challenging, e.g., these data exhibit unique challenges arising from heterogeneity and complexity. As heterogeneous data sources often do not follow data representation standards, integrating them into a common standard is another challenge~\cite{hasan2020knowledge}. An information extraction~(IE) method can recognize named entities classified as diseases or genetic from such unstructured data sources. IE in natural language processing~(NLP) involves \emph{named entity recognition}~(NER), \emph{relation extraction}~(RE), and \emph{entity linking}~(EL). 

Another challenge is semantic heterogeneity, which is further compounded by the flexibility of semi-structured data and various tagging methods applied. Semantic Web~(SW) technologies address data variety, by proposing graphs as a unifying data model, to which data can be mapped in a graph structure called knowledge graphs~(KGs). Nodes in a KG represent entities and edges represent binary relations between those entities~\cite{hogan2020knowledge}. A KG can be defined as $G=\{E, R,T\}$, where $G$ is a labelled and directed multi-graph, and $E, R, T$ are the sets of entities, relations, and triples, respectively and a triple can be represented as $(u,e,v) \in T$, where $u \in E$ is the head node, $v \in E$ is the tail node, and $e \in R$ is the edge connecting $u$ and $v$~\cite{hogan2020knowledge}. Another common way to represent extracted facts in Resource Description Framework~(RDF) format, where the linking structure of a KG forms a directed graph and triples are represented in the form of $(u,e,v)$ or $(subject,predicate,object)$. Facts containing these statements like~\textit{``TP53 is an oncogene"} or quantified statements~\textit{``oncogenes are responsible for cancer"} can be extracted from structured and unstructured sources and be integrated into a KG. 
A domain ontology is an important element for KG. Ontologies~(containing axioms or rules) and KBs that are increasingly adopted to address these challenges have great potential to support multidisciplinary cancer research~\cite{hogan2020knowledge}. Further, ontology-based NER and disambiguation help with the unambiguous identification of entities in heterogeneous data and the assertion of applicable named relationships that connect them. Moreover, ontology-based approaches for clinical decision support are getting wider adoption in the biomedical domain due to their explainability and reasoning capabilities~\cite{karim2022explainable}. 

Although reasoning over domain KGs enables consistency checking to recognize conflicting facts and deductive inferencing by revealing implicit knowledge from a set of facts~\cite{futia2020integration}, exploration, processing, and analyzing large-scale KGs is challenging. In this paper, we develop a domain KG biomarker discovery in cancer. We construct a domain ontology called OncoNet Ontology~(ONO). The KG is enriched by harmonizing the ONO and by extracting facts about concepts from scientific articles. Further, since the biomedical domain is evolving, where new findings often replace old ones, without using up-to-date domain knowledge there is a high chance an AI system exhibits concept drift. Therefore, it is important to validate the extracted facts with up-to-date domain knowledge~(e.g., domain experts). We finetuned the KG using large language models~(LLMs) based on more recent articles, and the ontology. 

\if false 
In this research, our contributions are as follows:
\begin{enumerate}
  \item We developed a knowledge graph (KG) specific to cancer, focusing on biomarker discovery.
  \item We introduced an ontology called OncoNet, which facilitates semantic reasoning for gene-disease validation tasks.
  \item We enriched the developed KG by harmonizing the introduced OncoNet with biomedical concepts extracted from scientific articles.
  \item To finetune the enriched KG, We leveraged the insights from large language models (LLMs) applied to recent articles and knowledge bases (KBs).
   
\end{enumerate}
\fi 

\section{Related work}\label{chapter_8:rw}
 Life sciences is one of the earliest adaptors of SW technologies~\cite{karim2022explainable}. Research initiatives in this area thus gradually employ semantic technologies such as KBs and domain-specific ontologies to build structured networks of interconnected knowledge~\cite{futia2020integration}. 
Subsequently, numerous research efforts from the scientific communities have focused on extracting semantic knowledge from diverse structured and unstructured data about cancer, followed by constructing large-scale KGs, by either manual annotation, crowd-sourcing~(e.g., DBpedia) or automatic extraction from unstructured data~(e.g., YAGO)~\cite{wang2015explicit} targeting specific use cases. The Bio2RDF and PubMed KGs~\cite{xu2020building} are two prominent KGs developed to accelerate bioinformatics research. Research efforts constantly expanding and evolving with more and more biomedical data. Another biomedical KG is built~\cite{alshahrani2017neuro} by harmonizing gene ontology~(GO), human phenotype ontology~(HPO), and disease ontology~(DO). Representation learning is then applied by combining symbolic logic and automated reasoning to generate node embeddings that are used for downstream tasks, e.g., link prediction, finding candidate genes of diseases, and protein-protein interactions. 

Another example is the prototype KG~\cite{hasan2020knowledge}, which is based on Louisiana Tumor Registry\footnote{\scriptsize{\url{https://sph.lsuhsc.edu/louisiana-tumor-registry/}}}. It provides scenario-specific querying, schema evolution for iterative analysis, and data visualization. Since this KG is built on a limited data setting, it does not contain comprehensive knowledge and facts about most cancer types. A large-scale breast cancer~(BRCA) KG is developed~\cite{hu2015semantic} by integrating: i) Dutch medical guidelines\footnote{\scriptsize{A semantic representation of Dutch medical guidelines.}} that contain conclusions and their evidence with UMLS and SNOMED CT medical terminologies, ii) genomic data of female patients, iii) clinical trials from official NCT website, iv) semantic annotations w.r.t eligibility criteria generated with XMedLan\footnote{\scriptsize{An NLP tool for biomedical texts.}}, and iv) selected medical publications from PubMed released by the Linked Life Data\footnote{\scriptsize{A semantic biomedical data integration platform.}}. 

Although these KGs are suitable for the exploration of the relation between knowledge and data sources about BRCA types, no comprehensive KGs have been developed targeting multiple cancer types to date. Further, the adoption of data-driven approaches has been hampered in many clinical settings by the lack of scalable computational methods~(e.g., models, KGs) that can deal with large-scale data that are heterogeneous, high dimensional, unstructured, and having high levels of uncertainty to perform in a reliable manner~\cite{phan2016integration}. 
Since the quality and consistency of the extracted facts in a KG are subject to the accuracy of the NER and RE, a domain expert requires reliable and up-to-date information. Further, since AI has become more widespread, the need for transparency of AI decisions has grown for ethical, legal, and safety reasons. This is more critical in healthcare, where AI may impact human lives~\cite{karim2022explainable}. Thus, there are many considerations for making AI-enabled healthcare applications more trustworthy~\cite{vollmer2020machine}. The field of explainable artificial intelligence~(XAI) aims to make AI systems more transparent and understandable to humans~\cite{karim2022explainable}. An interpretable ML model can reveal the factors that impact its outcomes. Model-specific and model-agnostic interpretable ML approaches have emerged~\cite{huang2023explainable}. 

Querying or IE from structured data is straightforward, while the same from unstructured data may be highly domain-specific and require efficient NLP methods. 
A concrete example is some selected cancer-specific biomarkers, e.g., some genes have both oncogenic and tumour-suppressor functionality called proto-oncogenes with tumour-suppressor function~(POTSF)~\cite{POSTF}. The majority of POTSF genes act as transcription factors or kinases and exhibit dual biological functions, e.g., both positively and negatively regulate transcription in cancer cells. Besides, specific cancer types like leukaemia are over-represented by POTSF genes, whereas some common cancer types like lung cancer are under-represented by them~\cite{POSTF}. Another type of gene called proto-oncogenes is a group of genes that cause normal cells to cancerous when they are mutated. Oncogenes result from the activation of proto-oncogenes, whereas a POSTF itself cause cancer when they are inactivated, e.g., TP53 is a POTSF and abnormalities of TP53 have been found in more than half of human cancers. Mutations in proto-oncogenes are dominant and a mutated version of a proto-oncogene is called an oncogene. 

\begin{figure*}[h]
	\centering
	\includegraphics[width=0.6\textwidth]{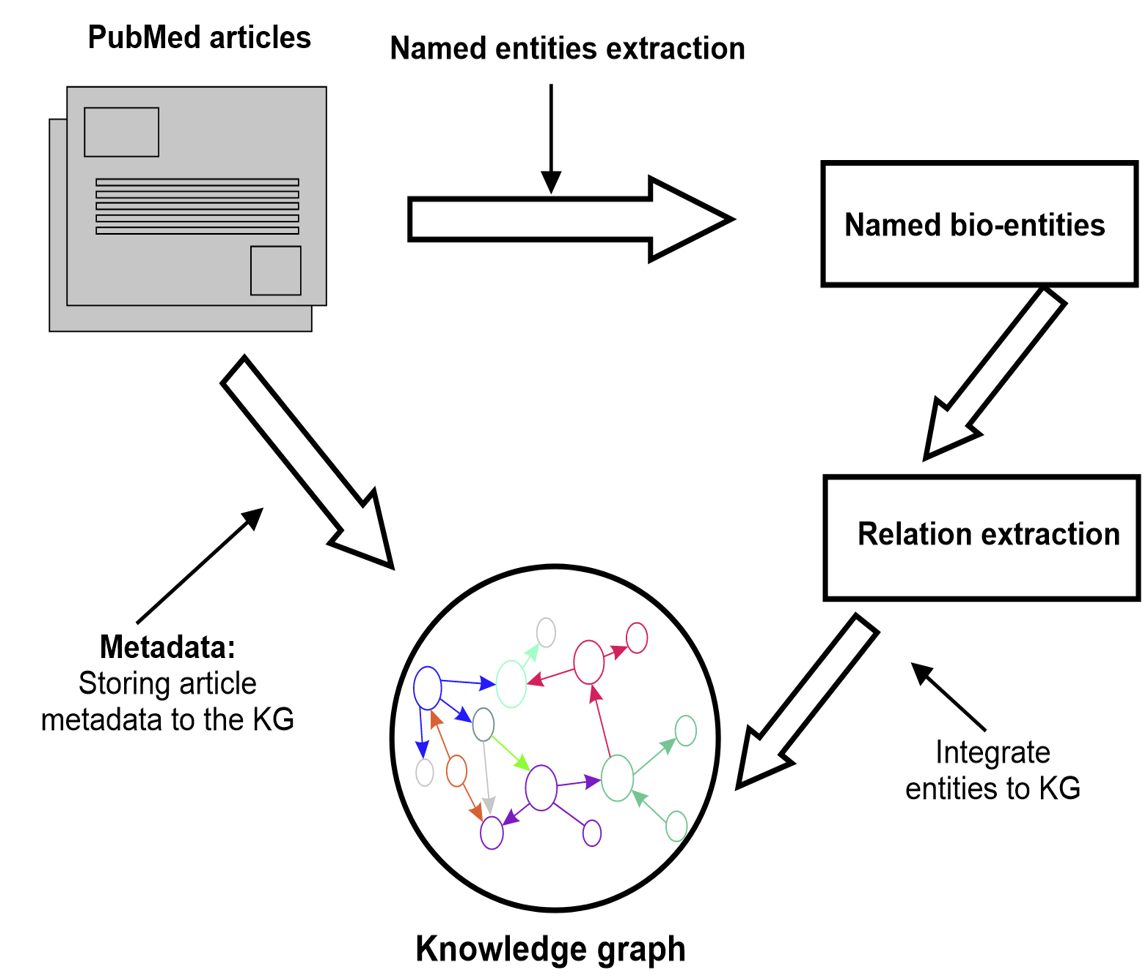}
	\caption{Schematic overview of the construction and building of our knowledge graph~(recreated based on KG building by~\cite{karim_phd_thesis_2022})} 
	\label{fig:kg_creation}
\end{figure*} 


Recently, transformer language models~(TLMs) have become the de facto standard for representation learning in NLP. BERT~\cite{devlin2018bert} utilizes a bidirectional attention mechanism and large-scale unsupervised corpora to obtain context-sensitive representations of words~\cite{xue2019fine}. Further, transformers are employed to predict entity labels from general texts~\cite{sun2020biomedical} as well as scientific texts~(e.g., SciBERT~\cite{Beltagy2019SciBERT}), where word representations are obtained from a pre-trained BERT model. BioBERT~\cite{lee2020biobert} is another variant of BERT that significantly outperforms DNN-based methods for bio-entity extraction.~Compared to transformers that benefit from abundant knowledge from pre-training and strong feature extraction capability, traditional conditional random field~(CRF) or neural networks~(DNNs)-based approaches such as Bidirectional Long Short Term Memory~(Bi-LSTM) or Gated Recurrent Unit~(GRU) cells have a lower generalization performance~\cite{xue2019fine}. These make TLMs of great potential for a variety of downstream NLP tasks. 

Large language models~(LLMs) such as GPT-4, LaMDA, and PaLM are expensive to train or fine-tune. LLMs pre-trained on large corpora inherently learn knowledge from a large text corpus during their pre-training plus open domain knowledge or new text provided through the prompts. Pre-trained LLM models calculate the probability of a sequence of words in a text, $T=\left(w_1, w_2, \ldots, w_L\right)$ is formulated as $p(T)=p\left(w_1\right) p\left(w_2 \mid w_1\right) \ldots p\left(w_L \mid w_1, w_2, \ldots, w_{L-1}\right)$, where $L$ is the input length~\cite{wang2023clinicalgpt}. 

\section{Methods}\label{chapter_8:mm}
\begin{figure*}
	\centering
	\includegraphics[width=\textwidth]{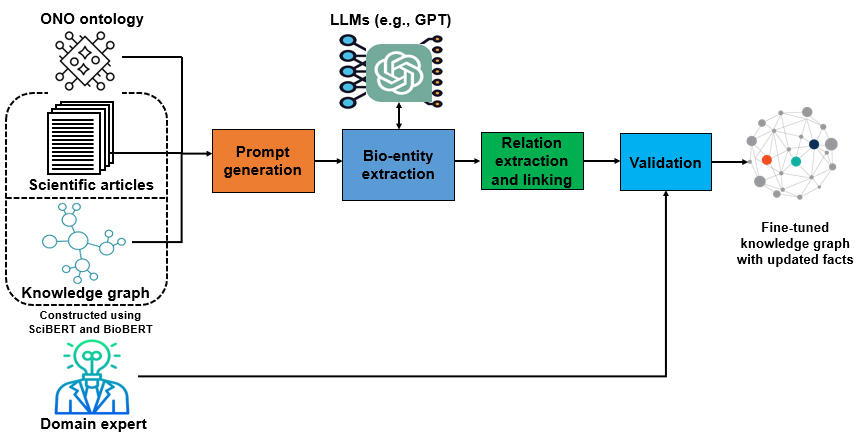}
	\caption{Fine-tuning and validation
 of the KG using LLMs~(based on 
 Text2KGBench by Mihindukulasooriya et al.~\cite{text2kgbench})} 
	\label{fig:llm_to_kgs}
\end{figure*}

First, we construct our ONO ontology, which is then used to enrich the KG along with additional controlled vocabularies and facts from scientific articles. Then, we construct the KG by integrating cancer-related knowledge from different sources. Finally, we use LLMs to fine-tune the KG, as shown in \cref{fig:llm_to_kgs}. 

\subsection{Ontology modelling}
The ONO is developed by reusing existing ontologies containing annotations about diseases, genes, concepts, and biological processes. We consider Cancer, Biomarker, and Feature classes as basis~(controlled vocabularies\footnote{\scriptsize{Concepts and their relations.}} and facts\footnote{\scriptsize{Factual knowledge from the literature.}}) for ONO: 

\begin{itemize}
\setlength\itemsep{0em}
    \item \texttt{ono:Cancer} class is a conceptual umbrella term that contains 33 cancer types.
    Entities are either biological or cancer-specific domain terms. 
    \item \texttt{ono:Biomarker} class is a conceptual umbrella term that contains 660 genes that are responsible for different types of cancer~(ref.  \texttt{ono:Cancer} class). A gene may be responsible for single or multiple types of cancer, e.g., TP53 is responsible for breast~(BRCA), ovarian~(OV), medulloblastoma~(MED), and prostate~(PRAD) cancer. For \emph{geneType}, `Oncogene', `Protein-coding', `POTFS' are possible types; for \emph{evidenceType}, `PubMed', `MeSH', `CancerIndex' are possible sources; for \emph{hasSignificance} a gene can have one of `HIGH', `MEDIUM', `LOW' levels of significance; \emph{crossResponsibility} can have one or more than one of `ACC', `BLCA', `BRCA', `CESC', `CHOL', `COAD', `DLBC', `ESCA', `GBM', `HNSC', `KICH', `KIRC', `KIRP', `LAML', `LGG', `LIHC', `LUAD', `LUSC', `MESO', `OV', `PAAD', `PCPG', `PRAD', `READ', `SARC', `SKCM', `STAD', `TGCT', `THCA', `THYM', `UCEC', `UCS', `UVM' 33-cancer types; for \emph{hasCitations}, there is at least 1 article indexed in `PubMed', `MeSH', `CancerIndex'. 
    \item The \texttt{ono:Feature} class characterizes additional information about the genes~(defined in \texttt{ono:Biomarker} class) such as biomarker types~(i.e., oncogenes, protein-coding, and POTSF), degree of significance w.r.t certain cancer types~(i.e., high, medium, and low), and source of evidence~(i.e., PubMed, CancerIndex, MeSH). For example, TP53 is highly significantly mutated in the breast~(BRCA) and ovarian~(OV) cancer types, significantly mutated in the prostate~(PRAD) cancer type, and nearly significantly mutated in medulloblastoma~(MED) cancer types. These findings are evidently found in scientific articles cited in PubMed, CancerIndex, and MeSH. 
\end{itemize}

\begin{figure*}[h]
	\centering
	\includegraphics[width=0.65\textwidth]{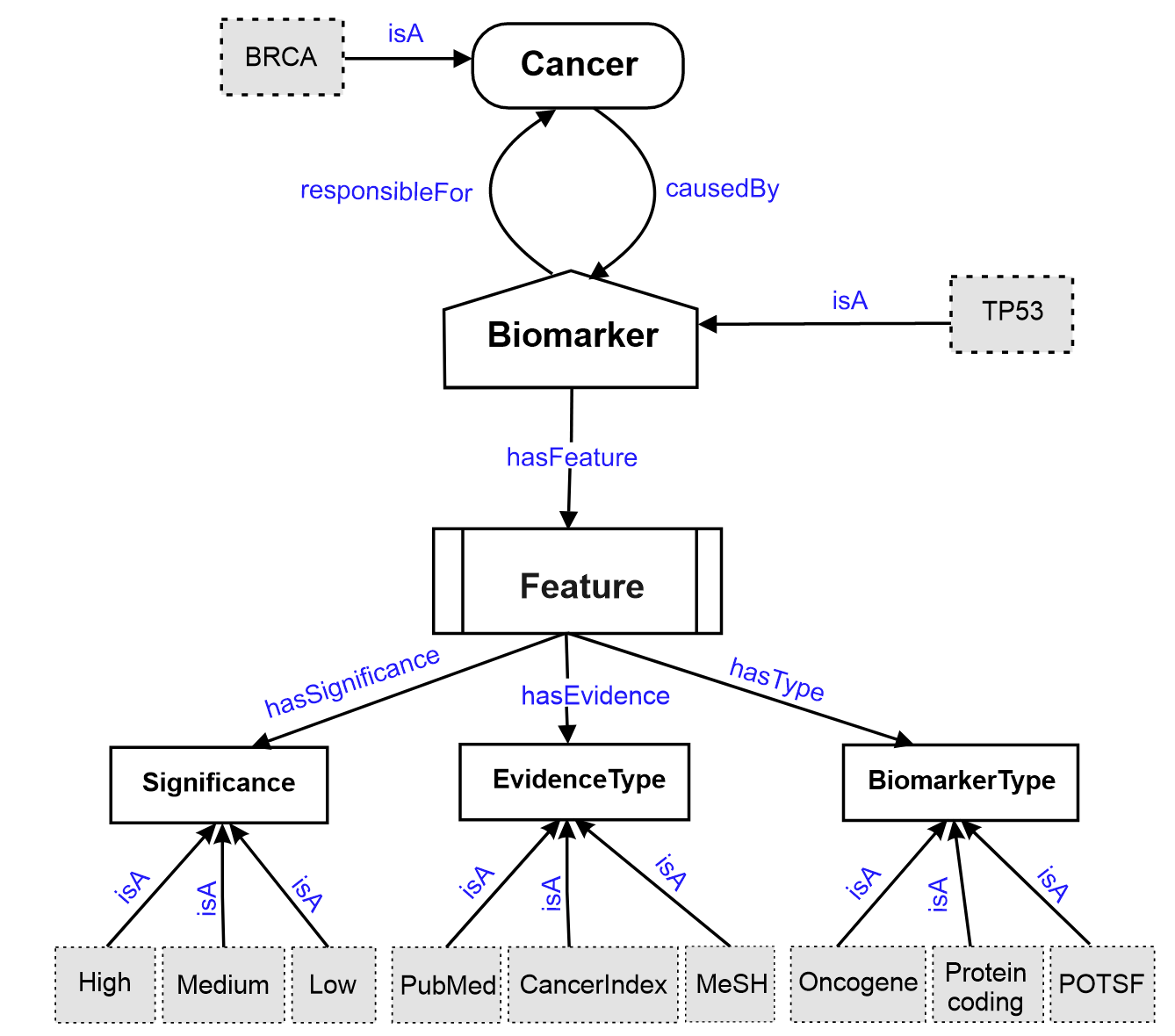}
	\caption[Cancer, Biomarker, and Feature classes and their properties that relate them]{Cancer, Biomarker, and Feature classes and their properties. Responsibility properties are shown as predicates, while dashed grey boxes signify properties of entities belonging to each class~\cite{karim_phd_thesis_2022}} 
	\label{fig:3classes_full}
\end{figure*}

We used ontology of genes and genomes~(OGG)\footnote{\scriptsize{\url{https://bioportal.bioontology.org/ontologies/OGG}}} and human disease ontology~(DOID)\footnote{\scriptsize{\url{https://bioportal.bioontology.org/ontologies/DOID}}} to inherit metadata about entities concerning genes and genomes, and human diseases, respectively. \Cref{fig:3classes_full} shows how classes relate to each other via properties, while \cref{fig:tp53_v2} shows how a biomarker~(e.g., TP53) is conceptualized via different classes.
We inherit semantic concepts about entities~(e.g., annotations and metadata) from GO, HPO, DO, breast cancer ontology~(BCO), cancer genetics web, TumorPortal\footnote{\scriptsize{\url{www.tumorportal.org}}}, and cancer genetic website\footnote{\scriptsize{\url{www.cancer-genetics.org/}}}. This reflects the structural aspects of the connection between biomarkers~(e.g., oncogenes, POTSF, ProteinCoding) and cancer types~(e.g. breast cancer). 

\subsubsection{Instance mapping} 
\begin{figure*}[h]
	\centering
	\includegraphics[width=\textwidth]{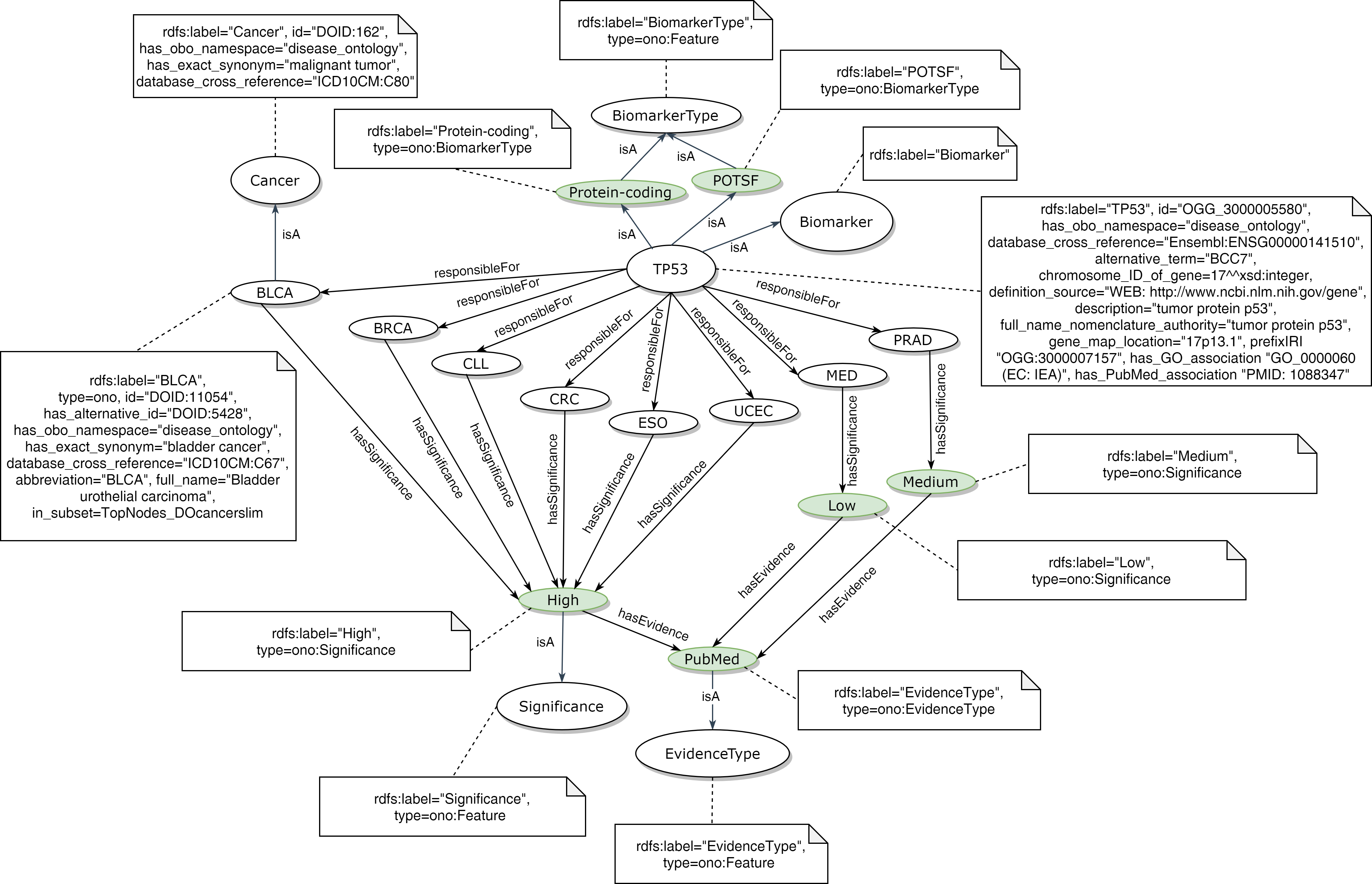}
	\caption[Properties and relevance of biomarkers w.r.t different types of cancer]{Properties of TP53 biomarker, showing how it is related to different types of cancer~\cite{karim_phd_thesis_2022}}
	\label{fig:tp53_v2}
	\vspace{-2mm}
\end{figure*}

DO provides consistent and reusable descriptions of human disease terms, phenotype characteristics, and medical vocabulary disease concepts. DO semantically integrates disease and medical vocabularies via cross-mapping of DO terms from MeSH, ICD, NCI, SNOMED, and OMIM. Besides, DO provides integrated information from several data sources and analysis of the literature using data from PubMed and CancerIndex.org. Therefore, we map all protein identifiers to Entrez genes and are used to represent genes, proteins, and other biological entities from the DO. 
Biological entities and classes from biomedical ontologies are two distinct types of entities. Each instance $f_i$ in the KG is assigned a unique IRI. Therefore, we treat biological entities like types of genes, proteins, and diseases as instances. Classes from DO are treated as instances. The general structure of the ontology is shown in \cref{fig:tp53_v2}, depicting hierarchical relations between different classes and subclasses. 
Ontology-based annotations are expressed by asserting a relation between instances like a gene, cancer type, or an instance of a class\footnote{\scriptsize{E.g., gene $AKT1$ has GO association \texttt{GO:0000060} by two axioms \texttt{hasGOAssociation(AKT1,$f_1$)} and \texttt{instanceOf($f_1$, GO:0000060)}: $AKT1$ and $f_1$ are instances, \texttt{GO:0000060} is a class \url{http://purl.obolibrary.org/obo/GO\_0000060} in GO, \texttt{hasGOAssociation} is an object property, and \texttt{instanceOf} is a \texttt{rdf:type} property.}}.

\subsection{Knowledge graph enrichment} 
 Annotations in the ONO are already rich enough to facilitate complex QA and reasoning. However, the ontology alone is not enough as it may contain incomplete knowledge. Therefore, additional facts~(e.g., some genes exhibit both oncogenic and tumour-suppressor characteristics, e.g., BRCA1, CAMTA1, CBFA2T3, CDX2, CREB3L1, CREBBP, DDB2, DNMT1, DNMT3A, ETV6, EZH2, FOXA1, FOXL2, FOXO1, FOXO3, FOXO4, FOXP1, FUS, IRF4, KLF4, KLF5, NCOA4, NOTCH1, NOTCH2, NOTCH3, NPM1, NR4A3, PAX5, PML, PPARG, RB1, RUNX1, SMAD4, STAT3, TCF3, TCF7L2, TP53, TP63, TRIM24, WT, ZBTB16, BCR, CHEK2, EPHA1, EPHA3, EPHB4, FLT3, MAP2K4, MAP3K4, MST1R, NTRK3, PRKAR1A, PRKCB, SYK, ARHGEF12, BCL10, BRCA2, CBL, CDC73, CDH11, CDKN1B, DCC, DDX3×, DICER1, FAS, FAT1, GPC3, IDH1, IKZF2, LIFR, NF2, NUP98, PHF6, PTPN1, PTPN11, RHOA, RHOB, SH2B3, SLC9A3R1, SOCS1, SPOP, SUZ12, WHSC1L1 are POTSF genes~\cite{POSTF}) from more recent articles need to be integrated. 
 
 To mitigate concept drift and enrich the KG, cancer-specific articles are collected from PubMed, 
where only recent and highly cited articles are considered. Pre-processing involves tokenisation, part-of-speech tagging, stemming, dependency parsing, word disambiguation, and linking words. We employ BERT-based information extraction, where BioBERT and SciBERT are finetuned with recent articles. Our hypothesis is that in learning to recover masked tokens, these NER models form a representational topology of cancer-specific articles that will help outperform CRF and Bi-LSTM for NER, relationship extraction, and multi-type normalization. 

\subsubsection{Named entity recognition}  
bio-entity extraction module serves two tasks: i) NER that recognizes the named entities in PubMed abstracts based on BioBERT~\cite{lee2020biobert} and SciBERT~\cite{Beltagy2019SciBERT} models, ii) EL to link extracted named entities, and iii) RE. NER is about recognizing domain-specific proper nouns in a biomedical corpus, e.g., for the sentence: \texttt{The main proteins regulating cell-cycle are BAD and PNCa, which have downstream effects}, proteins, cell-cycle, BAD, and PNCa are 4 proper nouns. Another example is \texttt{TP53 and FAS are the top two POTSF genes in terms of the number of associated cancer types, which are associated with 34 and 15 cancer types}, where a NER model is able to identify that TP53, FAS, POTSF genes and cancer are the named entities. Recent NER approaches extract such named entities based on learning frameworks that leverage lexical features such as POS tags, and dependency parse trees~\cite{hogan2020knowledge}. 

\begin{figure*}
	\centering
	\includegraphics[width=\textwidth]{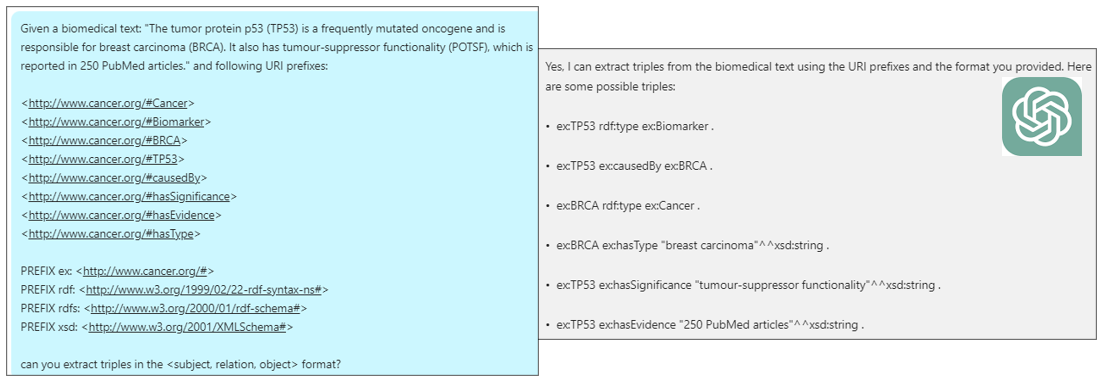}
	\caption{An example prompt to generate triples from the given text and ontology as input to ChatGPT} 
	\label{fig:propt_llms}
\end{figure*} 

BERT was originally pre-trained on English Wikipedia, news, and book corpus. Therefore, it requires fine-tuning on biomedical texts containing domain-specific proper nouns and terms. Inspired by the success of BioBERT and SciBERT at biomedical NER tasks~\cite{kim2019neural}, we fine-tuned them to perform NER. 
BioBERT and SciBERT are first initialized with a case-sensitive version of BERT, followed by fine-tuning them on PubMed Central full-text articles collected based on inclusion and exclusion criteria. While fine-tuning BioBERT and SciBERT, WordPiece tokenization is used in which any new words are represented by frequent sub-words. SciBERT and BioBERT extract different entity types, where an entity or two with frequently occurring token interaction is marked with more than one entity type span. Then, based on probability distribution, we choose the correct entity when they were tagged with more than two types w.r.t probability-based decision rules~\cite{kim2019neural}. These two NER models can predict 7 tags: IOB, X, CLS, SEP, and PAD. 

\subsubsection{Entity linking and relation extraction} for the entities present in the KG, we link given mentions to these nodes, involving entity disambiguation, e.g., multiple ways exist to mention the same entity~(e.g., \texttt{TP53} and \texttt{Li-Fraumeni syndrome} are anonymously used). Further, multi-type normalization is performed to assign unique IDs to extracted bio entities. For the RE, we consider both binary and n-ary relation types, in a closed-world setting between gene/protein and disease relation types. We utilize the sentence classifier of the BERT-case. BERT uses a [CLS] token for the relation classification. Similar to BioBERT-based RE, sentence classification is performed using a single output layer based on a [CLS] token representation from BERT. The target named entities in a sentence are anonymized using pre-defined @GENE\$ and @DISEASE\$.  

\subsubsection{Integrating extracted triples into KG} possible sub-classes of \emph{Significance} is defined as HIGH, MEDIUM, and LOW based on the annotations provided in the TumorPortal\footnote{\scriptsize{\url{www.tumorportal.org/tumor_types?ttype=PanCan}}}. Each entity belonging to the biomarker class is annotated with the properties from OGG ontology. Besides, the number of  articles is included as evidence associated with certain genes. We included those 33 cancer types of interest in the Cancer class, which are labelled with their well-known abbreviations, e.g., BRCA for breast cancer. For the entities in the Cancer class, object properties from the DOID ontology are inherited. The \emph{Feature} class connects the disease and responsible genes. The significance degree of a particular gene is included w.r.t a disease. Besides, biomarker and evidence-type information are also included, where PubMed, MeSH, and CancerIndex are considered as the source of the evidence. These features are used to indicate the relation between entities from different classes, which in turn annotations allow the rules generation. Since the genes can be categorized as \emph{Oncogene}, \emph{ProteinCoding}, and \emph{POTSF}, we considered these as the subclasses of \emph{BiomarkerType}. 

Links between extracted lexical terms from the source text and the concepts from the ontology are defined. Then, the context of the terms is analyzed to determine appropriate disambiguation, before assigning them the correct concept.~Finally, attribute-value pairs are identified, which involves the identification of a subject, mapping it to a semantic class, and using the predicate and object as the attribute name and value, respectively. We use a semantic lexicon to integrate new facts into our KG. Our BioBERT and SciBERT-based instance lexicon extractor uses rules from the ontology to enrich the instance information to create RDF triples, where each triple forms a connected component of a sentence, e.g., for input text, \textit{``TP53 is responsible for a disease called Breast Cancer. TP53 has POTSF functionality, which is mentioned in numerous PubMed articles.''}, \emph{TP53}, \emph{disease}, \emph{Breast Cancer}, \emph{POTSF}, and \emph{PubMed} are named entities. This yields following triples: \texttt{(TP53, causes, Breast Cancer)}, \texttt{(TP53, hasType, POTSF)}, \texttt{(Breast Cancer, isA, Disease)}, \texttt{(POTSF, hasEvidence, PubMed)}. 

\begin{figure*}
	\centering
		\includegraphics[width=\textwidth]{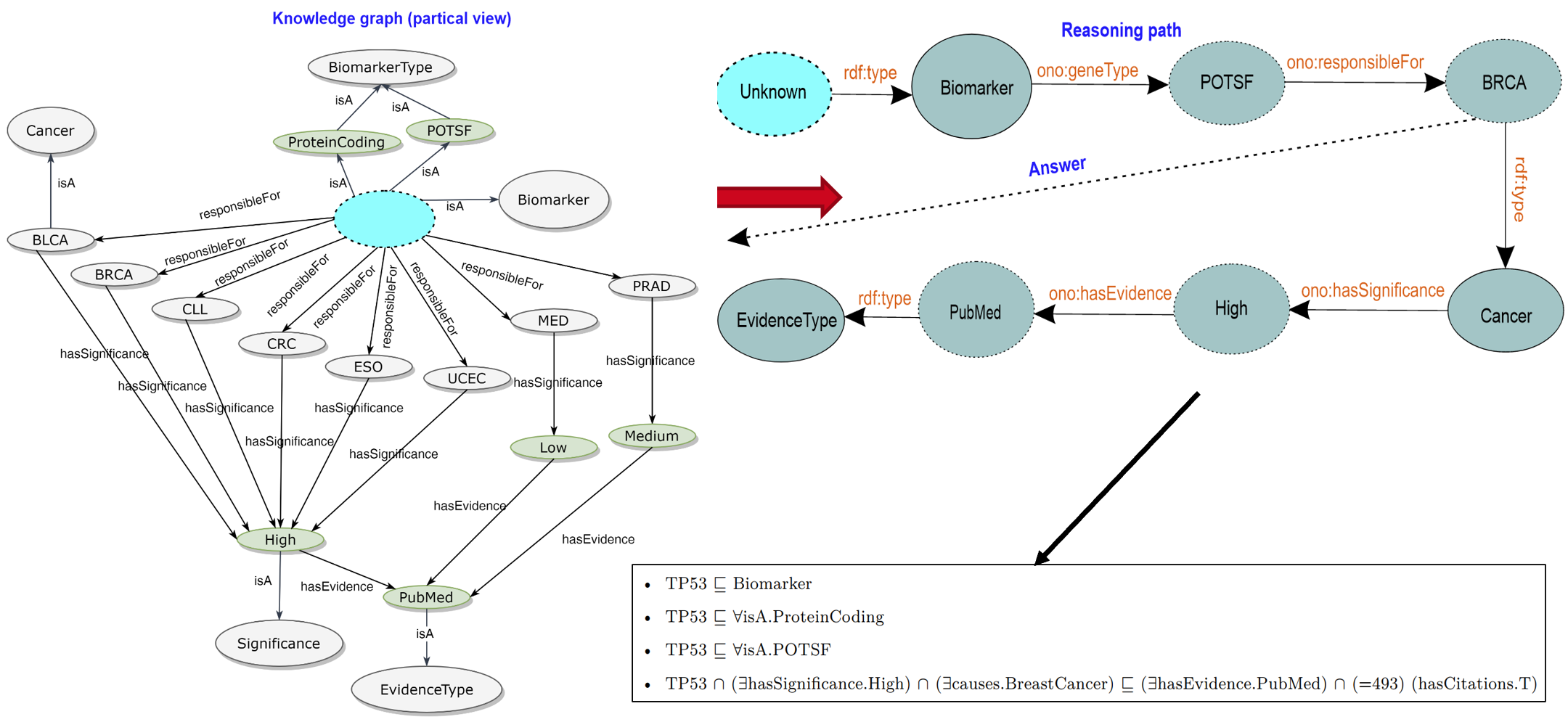}
        \label{fig:ner_lrp}
	\caption{Example of question answering to infer knowledge about named bio-entities~\cite{karim_phd_thesis_2022}}
\end{figure*}

\subsection{Fine-tuning KG with LLMs}
 Assuming our KG may have potentially inconsistent or outdated facts, we use an LLM not only to extract up-to-date knowledge but also to identify such inconsistencies or incompleteness. We employ the instruction-tuning approach with supervised fine-tuning~(SFT)~\cite{wang2023clinicalgpt} and Text2KGBench~\cite{text2kgbench}. Unlike training LLMs from scratch, we leverage their capabilities for IE with in-context learning in prompts. As shown in \cref{fig:propt_llms}, an LLM is provided with our ontology, the KG, and reference\footnote{\scriptsize{The full-text corpora with guided URIs and prefixes.}} about very recent articles that may not have been used during the fine-tuning of BioBERT or SciBERT. Thus, our goal is to generate triples from the input texts guided by the ontology, followed by comparing them with existing triples in the KG. For a given input prompt $I=w_{1: m}$, we instruct the ChatGPT model $f_\theta$ to generate a response $R=v_{1: n}$ by optimizing the likelihood $f_\theta(R \mid I)=p_\theta\left(v_{1: n} \mid w_{1: m}\right)$~\cite{wang2023clinicalgpt}, where $n$ and $m$ are the lengths of response and input prompt, respectively. The response is then represented as RDF triples for our KG. Finally, we perform automated validation of facts via semantic reasoner. 

\subsection{Quality assessment of knowledge graph}
Regardless of the data sources, the initial KG will usually be incomplete and may  contain duplicates, inconsistencies, or even incorrect statements – especially when a KG is constructed based on multiple sources~\cite{hogan2020knowledge}. Therefore, once the KG is constructed and enriched with additional facts from external sources, we follow another crucial step, which is the quality assessment of the facts in the KG to assess the fitness of the purpose w.r.t availability, completeness, conciseness, interlinking, performance, and relevancy dimensions, using SANSA-linked data quality assessment metrics~\cite{sejdiu2019scalable}. 

\section{Evaluation Results}\label{chapter_8:results}
In this section, we report some experiment results. 

\subsection{Experiment setup}
 Each \textit{BERT}-based NER model is fine-tuned for 20 epochs~(the training set is shuffled for each epoch, where the maximum input length is set to $256$), where the Adam optimizer is used to optimize the loss with the scheduled learning rate~(in which the initial learning rate is set to $2 e^{-5}$ with gradient clipping applied). Results produced through random search and 5-fold cross-validation tests are reported. To evaluate the effectiveness of domain-specific potentials of transformer-based approach for information extraction, a comparative analysis with the well-known CRF and Bi-LSTM architectures is provided. Further, some benchmark queries for selected questions are provided in human language~(NLQ) and DLx query~(DLQ) formats, where the DLx rules are generated using Pellet, ELK, and HermiT reasoners. Prot{\'e}g{\'e} 5.5.0 is used for the reasoning, w.r.t DLx querying. However, we report the rules generated with Pellet reasoner only\footnote{\scriptsize{ELK does not support object all values from axioms, while HermiT gave some inconsistencies, yielding a different number of rules.}}. The inferred rules based on reasoning are expressed in DLx format. Finally, generated rules are interpreted, followed by decision reasoning with the rules. 

\subsection{Analysis of information extraction}
 With \emph{exact-match evaluation}, a named entity is considered correctly recognized by the NER model only if both boundaries and type match the ground truth. To validate the knowledge extraction from scientific articles using BioBERT and SciBERT, the performance of the entity extraction is evaluated in terms of entity-level precision and recall. While precision measures the ability of a NER system to present only correct entities, recall measures the ability of the NER model to recognize all entities in our corpus. We report  precision, recall, and F1 scores in \cref{table:ner_results}, with the best scores in bold and the second-best scores underlined. Since both high precision and high recall are desirable, an F1 score was not shown. Results for BERT and BiLSTM-CRF models are provided as two different baselines. 

\begin{table}[h!]
    \centering
    \caption{Performance of entity recognition and linking}
    \label{table:ner_results}
    \scriptsize{
    \begin{tabular}{l|l|l|l|l} 
    \hline
        & \multicolumn{2}{c|}{Entity extraction} & \multicolumn{2}{c}{Normalization} \\ 
        \hline
        \textbf{Model} & \textbf{Precision} & \textbf{Recall} & \textbf{Precision} & \textbf{Recall} \\ 
        \hline
        BiLSTM-CRF  & 83.13\% & 82.19\% & 84.57\% & 83.29\% \\ 
        \cline{1-3}\cline{4-5}
        BERT & 87.25\% & 86.65\% & 88.34\% & 87.31\% \\ 
        \hline
        SciBERT & \underline{89.35\%} &  \underline{88.55\%} &  \underline{90.12\%} &  \underline{89.37\%} \\
        \hline
        BioBERT &\textbf{ 91.36\%} & \textbf{90.75\%} & \textbf{91.32}\% & \textbf{91.43\%} \\
        \hline
    \end{tabular}}
\end{table}

BioBERT obtained the highest F1-score in recognizing Genes/Proteins and diseases, which is about 2\% better than SciBERT. BioBERT significantly outperformed the BiLSTM-CRF model by $7.85 \%$ w.r.t F1-score. Being fine-tuned on domain-specific articles, SciBERT also outperformed the BiLSTM-CRF by $5.39 \%$ in terms of the F1-score, on average. BERT, which is pre-trained on general domain corpus was highly effective. On average, BioBERT and SciBERT outperformed BERT by 3.25\% in terms of F1-score. 

\begin{figure*}[h]
	\centering
	\includegraphics[width=\textwidth]{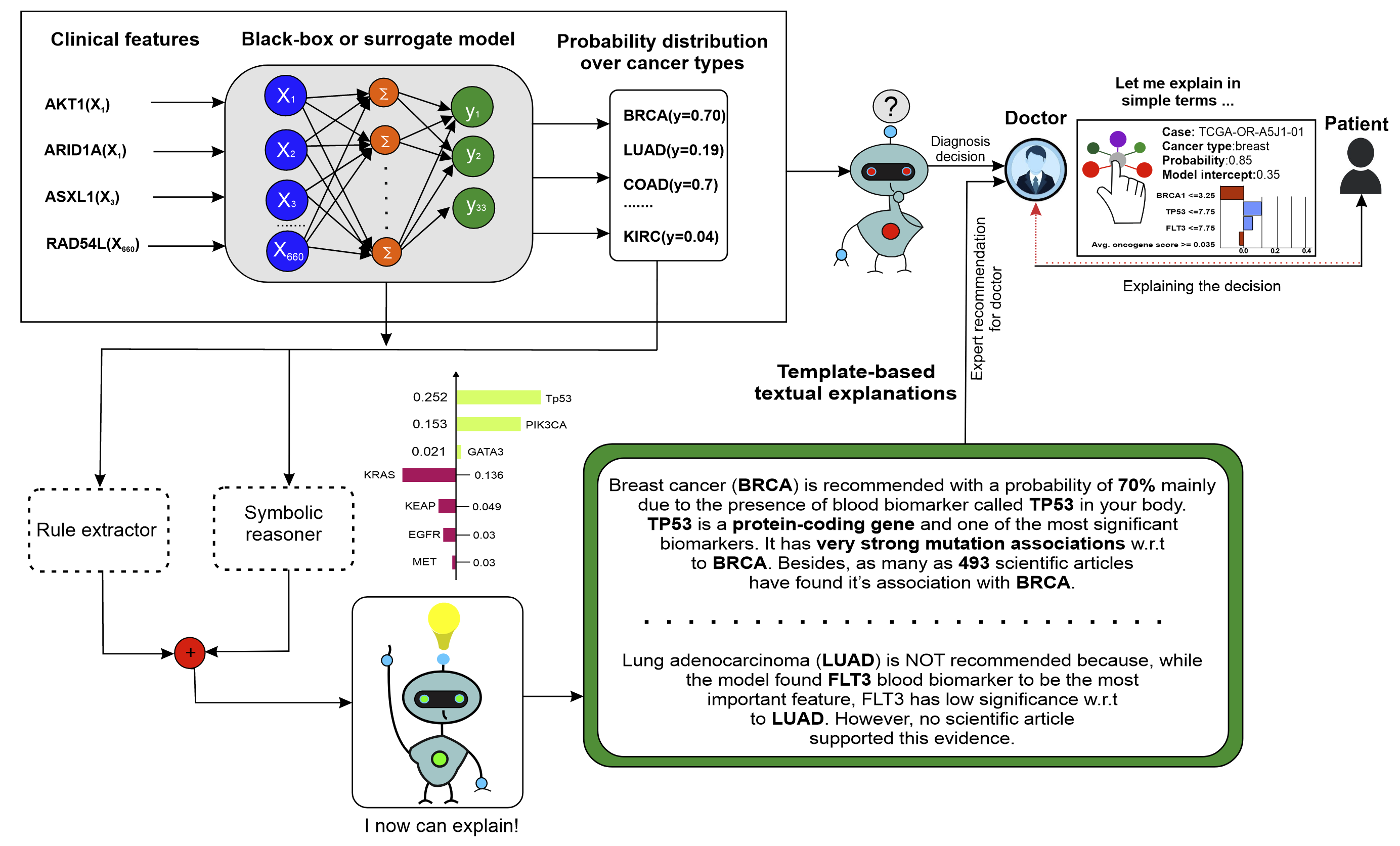}
	\caption{Decision reasoning on facts from a domain-knowledge graph~\cite{karim2022explainable}}
    \label{fig:decision_reasoning_with_rules}
\end{figure*}

\subsection{Using the knowledge graph}
 Explanations serve as a bridge between humans and AI systems~\cite{guidotti2018survey}, where the AI itself could benefit from external knowledge to support domain experts in understanding why the algorithms came up with certain results. 
We foresee the benefits for both researchers~(e.g., bioinformatician and SW researchers to leverage the interactive explanation via querying and question answering~(QA) and medical doctors~(e.g., oncologists can validate diagnosis decisions) based on explicit knowledge from the KG. For example, the answer to NLQ: ``List of all biomarkers that are classified POTSF and responsible for breast cancer" for the DLQ ``Biomarker \textcolor{red}{and} causes \textcolor{blue}{some} BRCA \textcolor{red}{and} isA \textcolor{green}{only} POTSF". Since our KG can be viewed as discrete symbolic representations of knowledge, reasoning over it would leverage the symbolic technique\footnote{\scriptsize{Inference rules~(IRs) are a straightforward way to provide automated access to deductive knowledge~\cite{hogan2020knowledge}}. An IR encodes IF-THEN-style consequences: $p \rightarrow y$, where body and head follow graph patterns in a KG.}. These would help QA for NLQs, e.g., \emph{``Which POTSF biomarker is highly responsible for breast Carcinoma and has PubMed evidence?"}, the semantic reasoner follows a logical reasoning path to reason about the concept \emph{`unknown'} and how it is related to other concepts. 

Suppose an ML model provides diagnosis decisions for a cancer patient by identifying statistically significant biomarkers based on SHAP~\cite{huang2023explainable}. However, it is not evident that those biomarkers are biologically significant. The findings, however, can be validated based on facts presented in the KG. The integration of an ML model with a knowledge-based system would provide human operators with both reasoning, QA, and validating the predictions\footnote{\scriptsize{\emph{Neuro-symbolic AI} that combines connectionist- and symbolic AI paradigms.}}. Assuming our KG is complete in the closed-world assumption, a doctor with their expertise and by combining the facts from KG could explain the decision with additional interpretation~(e.g., biomarkers and their relevance w.r.t specific cancer types), as shown in \cref{fig:decision_reasoning_with_rules}. 

\section{Conclusion}\label{chapter_8:conclusion}
 In this paper, we constructed a domain KG that can leverage cancer-specific biomarker discovery and interactive QA over it, while our domain ONO enables semantic reasoning for the validation of gene-disease relations in symbolic contexts. To mitigate the concept drift, we fine-tuned the KG using LLMs based on more recent articles and KBs. Our approach suggests that LLMs indeed help extract new facts and knowledge from scientific literature and help keep the kG up-to-date. 

We would like to outline potential limitations of our approach that leave a lot of improvement possibilities:~
First, the expressiveness of DLx is not tested against incompleteness and inconsistency. Second, we could use the fully inferred, deductively closed KG to perform representation learning, e.g., embeddings of nodes and relations. Such KG embeddings on the deductively closed graph would have the advantage that not only asserted axioms will be taken into consideration, but representations can include inferred knowledge that is not present explicitly in KG~\cite{alshahrani2017neuro}. 
Since the full potential of an AI system can only be exploited by integrating both domain- and human expertise~\cite{karim2022explainable}, we envision developing a neuro-symbolic AI system similar to in \cref{fig:decision_reasoning_with_rules}, by combining both connectionist- and symbolic AI paradigms not only to make knowledge acquisition and exploration but also reasoning and explainability. 

\section*{Acknowledgments}
 This paper is based on PhD thesis~\cite{karim_phd_thesis_2022} by the first author and partially funded by the Federal Ministry of Education and Research~(BMBF) project WEST-AI under grant no. 01IS22094E.

\bibliographystyle{natbib}
\bibliography{references}

\begin{thebibliography}{}

\bibitem[Alshahrani {\em et~al.}(2017)Alshahrani, Khan, and Hoehndorf]{alshahrani2017neuro}
Alshahrani M, Khan M.~A, and Hoehndorf R (2017).
\newblock Neuro-symbolic representation learning on biological knowledge graphs.
\newblock {\em Bioinformatics\/}, {\bf 33}(17), 2723--2730.

\bibitem[Ballester(2021)Ballester]{ballester2021artificial}
Ballester P.~J (2021).
\newblock Artificial intelligence for the next generation of precision oncology.
\newblock {\em NPJ Precision Oncology\/}, {\bf 5}(1), 1--3.

\bibitem[Beltagy and Cohan(2019)Beltagy and Cohan]{Beltagy2019SciBERT}
Beltagy I and Cohan A (2019).
\newblock {SciBERT: Pretrained Language Model for Scientific Text}.
\newblock In {\em EMNLP\/}.

\bibitem[Devlin and Lee(2018)Devlin and Lee]{devlin2018bert}
Devlin J and Lee K. e.~a (2018).
\newblock {BERT}: Pre-training of deep bidirectional transformers for language understanding.
\newblock {\em arXiv:1810.04805\/}.

\bibitem[Futia and Vetr{\`o}(2020)Futia and Vetr{\`o}]{futia2020integration}
Futia G and Vetr{\`o} A (2020).
\newblock On the integration of knowledge graphs into deep learning models for a more comprehensible ai—three challenges for future research.
\newblock {\em Information\/}, {\bf 11}(2), 122.

\bibitem[Guidotti {\em et~al.}(2018)Guidotti, Monreale, Ruggieri, Turini, Giannotti, and Pedreschi]{guidotti2018survey}
Guidotti R, Monreale A, Ruggieri S, et~al. (2018).
\newblock A survey of methods for explaining black box models.
\newblock {\em ACM computing surveys (CSUR)\/}, {\bf 51}(5), 1--42.

\bibitem[Hasan {\em et~al.}(2020)Hasan, Rivera, Wu, Durbin, Christian, and Tourassi]{hasan2020knowledge}
Hasan S.~S, Rivera D, Wu X.-C, et~al. (2020).
\newblock Knowledge graph-enabled cancer data analytics.
\newblock {\em IEEE journal of biomedical and health informatics\/}, {\bf 24}(7), 1952--1967.

\bibitem[Hogan {\em et~al.}(2021)Hogan, Blomqvist, Cochez, Navigli, Neumaier, {\em et~al.}]{hogan2020knowledge}
Hogan A, Blomqvist E, Cochez M, et~al. (2021).
\newblock Knowledge graphs.
\newblock {\em ACM Computing Surveys (Csur)\/}, {\bf 54}(4), 1--37.

\bibitem[Hu {\em et~al.}(2015)Hu, Huang, and Gu]{hu2015semantic}
Hu Q, Huang Z, and Gu J (2015).
\newblock Semantic representation of evidence-based medical guidelines and its use cases.
\newblock {\em Wuhan University Journal of Natural Sciences\/}, {\bf 20}(5), 397--404.

\bibitem[Huang {\em et~al.}(2023)Huang, Suominen, Liu, Rice, Salomon, and Barnard]{huang2023explainable}
Huang W, Suominen H, Liu T, et~al. (2023).
\newblock Explainable discovery of disease biomarkers: The case of ovarian cancer to illustrate the best practice in machine learning and shapley analysis.
\newblock {\em Journal of Biomedical Informatics\/}, {\bf 141}, 104365.

\bibitem[Karim {\em et~al.}(2022)Karim, Beyan, Rebholz-Schuhmann, and Decker]{karim_phd_thesis_2022}
Karim R, Beyan O, Rebholz-Schuhmann D, and Decker S (2022).
\newblock {Interpreting Black-box Machine Learning Models with Decision Rules and Knowledge Graph Reasoning}.

\bibitem[Karim {\em et~al.}(2023)Karim, Islam, Beyan, Rebholz-Schuhmann, and Decker]{karim2022explainable}
Karim R, Islam T, Beyan, et~al. (2023).
\newblock {Explainable AI for Bioinformatics: Methods, Tools, and Applications}.
\newblock {\em Briefings in Bioinformatics\/}.

\bibitem[Kim {\em et~al.}(2019)Kim, Lee, Jeong, Sung, and Kang]{kim2019neural}
Kim D, Lee J, Jeong M, et~al. (2019).
\newblock A neural named entity recognition and multi-type normalization tool for biomedical text mining.
\newblock {\em IEEE Access\/}, {\bf 7}, 73729--73740.

\bibitem[Kitsios {\em et~al.}(2023)Kitsios, Kamariotou, and Talias]{kitsios2023recent}
Kitsios F, Kamariotou M, and Talias M.~A (2023).
\newblock Recent advances of artificial intelligence in healthcare: A systematic literature review.
\newblock {\em Applied Sciences\/}, {\bf 13}(13), 7479.

\bibitem[Lee {\em et~al.}(2020)Lee, Yoon, Kim, So, and Kang]{lee2020biobert}
Lee J, Yoon W, Kim S, et~al. (2020).
\newblock {BioBERT: A Pre-trained Biomedical Language Representation Model for Biomedical Text Mining}.
\newblock {\em Bioinformatics\/}, {\bf 36}(4), 1234--1240.

\bibitem[Mihindukulasooriya {\em et~al.}(2023)Mihindukulasooriya, Tiwari, and Lata]{text2kgbench}
Mihindukulasooriya N, Tiwari S, and Lata K (2023).
\newblock {Text2KGBench}: A benchmark for ontology-driven knowledge graph generation from text.
\newblock In {\em ISWC\/}, pages 247--265. Springer.

\bibitem[Phan and Wang(2016)Phan and Wang]{phan2016integration}
Phan J.~H and Wang M.~D (2016).
\newblock Integration of multi-modal biomedical data to predict cancer grade and patient survival.
\newblock In {\em IEEE-EMBS International Conference on Biomedical and Health Informatics~(BHI)\/}, pages 577--580. IEEE.

\bibitem[Sejdiu {\em et~al.}(2019)Sejdiu, Rula, Lehmann, and Jabeen]{sejdiu2019scalable}
Sejdiu G, Rula A, Lehmann J, and Jabeen H (2019).
\newblock A scalable framework for quality assessment of rdf datasets.
\newblock In {\em International Semantic Web Conference\/}, pages 261--276. Springer.

\bibitem[Shen and Wang(2018)Shen and Wang]{POSTF}
Shen L and Wang W (2018).
\newblock Double agents: genes with both oncogenic and tumor-suppressor functions.
\newblock {\em Oncogenesis\/}, {\bf 7}(3), 1--14.

\bibitem[Sun and Wang(2020)Sun and Wang]{sun2020biomedical}
Sun C and Wang J (2020).
\newblock {Biomedical Named Entity Recognition using BERT in the Machine Reading Comprehension Framework}.
\newblock {\em arXiv:2009.01560\/}.

\bibitem[Tran {\em et~al.}(2021)Tran, Kondrashova, Pearson, and Waddell]{tran2021deep}
Tran K.~A, Kondrashova O, Pearson J.~V, and Waddell N (2021).
\newblock Deep learning in cancer diagnosis, prognosis and treatment selection.
\newblock {\em Genome Medicine\/}, {\bf 13}(1), 1--17.

\bibitem[Vollmer {\em et~al.}(2020)Vollmer, Mateen, Bohner, Kir{\'a}ly, Cumbers, Jonas, McAllister, and Myles]{vollmer2020machine}
Vollmer S, Mateen B.~A, Bohner G, et~al. (2020).
\newblock Machine learning and artificial intelligence research for patient benefit: 20 critical questions on transparency, replicability, ethics, and effectiveness.
\newblock {\em bmj\/}, {\bf 368}.

\bibitem[Wang {\em et~al.}(2023)Wang, Yang, and Li]{wang2023clinicalgpt}
Wang G, Yang G, and Li X (2023).
\newblock {ClinicalGPT: Large Language Models Finetuned with Diverse Medical Data and Comprehensive Evaluation}.
\newblock {\em arXiv:2306.09968\/}.

\bibitem[Wang {\em et~al.}(2015)Wang, Wu, Shen, and Dick]{wang2015explicit}
Wang P, Wu Q, Shen C, and Dick A (2015).
\newblock Explicit knowledge-based reasoning for visual question answering.
\newblock {\em arXiv:1511.02570\/}.

\bibitem[Xu {\em et~al.}(2020)Xu, Kim, Song, Jeong, Kim, Kang, Rousseau, Li, Xu, Torvik, {\em et~al.}]{xu2020building}
Xu J, Kim S, Song M, et~al. (2020).
\newblock Building a pubmed knowledge graph.
\newblock {\em Scientific data\/}, {\bf 7}(1), 1--15.

\bibitem[Xue {\em et~al.}(2019)Xue, Zhou, and He]{xue2019fine}
Xue K, Zhou Y, and He P (2019).
\newblock {Fine-tuning BERT for Joint Entity and Relation Extraction in Chinese Medical Text}.
\newblock In {\em Conference on Bioinformatics and Biomedicine~(BIBM)\/}, pages 892--897. IEEE.

\end{thebibliography}

\end{document}